\pdfoutput=1

\documentclass[11pt]{article}
\usepackage{graphicx}
\usepackage[table,xcdraw,dvipsnames]{xcolor}
\usepackage{multirow}
\usepackage{enumitem}
\usepackage{adjustbox}
\usepackage{tcolorbox}
\usepackage{booktabs} 
\usepackage{hyperref}
\usepackage{array} 
\usepackage{makecell}

\usepackage{geometry} 

\usepackage{longtable} 

\definecolor{headergray}{gray}{0.8}
\definecolor{sectionblue}{RGB}{173,216,230}

\usepackage{ACL2023}

\usepackage{times}
\usepackage{latexsym}
\usepackage{amsfonts}
\usepackage[T1]{fontenc}

\usepackage[utf8]{inputenc}

\usepackage{microtype}

\usepackage{inconsolata}

\title{

Legal Judgment Reimagined: PredEx and the Rise of Intelligent AI Interpretation in Indian Courts

}
\author{Shubham Kumar Nigam$^{1* \dagger}$ \qquad Anurag Sharma$^{2*}$ \qquad Danush Khanna$^{3*}$ \\ 
\textbf{Noel Shallum}$^{4}$ \qquad \textbf{Kripabandhu Ghosh}$^{2}$ \qquad \textbf{Arnab Bhattacharya}$^{1}$\\
$^{1}$ Indian Institute of Technology Kanpur, India \quad
$^{2}$  IISER Kolkata, India \\
$^{3}$ Manipal University Jaipur, India \quad
$^{4}$ Symbiosis Law School Pune, India \\
\texttt{sknigam@cse.iitk.ac.in,}  \quad \texttt{as19ms159@iiserkol.ac.in,} \\
\texttt{danush.229310455@muj.manipal.edu,}\quad
\texttt{kripaghosh@iiserkol.ac.in,} \\
\texttt{noelshallum@gmail.com,} \quad
\texttt{arnabb@cse.iitk.ac.in}
}

\begin{document}
\maketitle

{
\renewcommand{\thefootnote}{$*$}
\footnotetext{These authors contributed equally to this work}
\renewcommand{\thefootnote}{$\dagger$}
\footnotetext{Corresponding author}
\renewcommand{\thefootnote}{\arabic{footnote}}
}

\begin{abstract}
In the era of Large Language Models (LLMs), predicting judicial outcomes poses significant challenges due to the complexity of legal proceedings and the scarcity of expert-annotated datasets. Addressing this, we introduce \textbf{Pred}iction with \textbf{Ex}planation (\texttt{PredEx}), the largest expert-annotated dataset for legal judgment prediction and explanation in the Indian context, featuring over 15,000 annotations. This groundbreaking corpus significantly enhances the training and evaluation of AI models in legal analysis, with innovations including the application of instruction tuning to LLMs. This method has markedly improved the predictive accuracy and explanatory depth of these models for legal judgments. We employed various transformer-based models, tailored for both general and Indian legal contexts. Through rigorous lexical, semantic, and expert assessments, our models effectively leverage \texttt{PredEx} to provide precise predictions and meaningful explanations, establishing it as a valuable benchmark for both the legal profession and the NLP community.
\end{abstract}

\section{Introduction}
\begin{table*}[t]
\centering
\resizebox{\textwidth}{!}{%
\begin{tabular}{|c|c|c|c|c|c|l|l|}
\hline
\textbf{Corpus} &
  \textbf{Language} &
  \textbf{Jurisdiction} &
  \textbf{\# of Cases} &
  \textbf{\begin{tabular}[c]{@{}c@{}}\# of Human \\ annotated Docs\end{tabular}} &
  \begin{tabular}[c]{@{}c@{}}\textbf{Avg \# of} \\ \textbf{Tokens}\end{tabular} &
  \multicolumn{1}{c|}{\textbf{\begin{tabular}[c]{@{}c@{}}Annotated LJP Subtasks\\ (\# of labels w.r.t Subtask)\end{tabular}}} &
  \multicolumn{1}{c|}{\textbf{Additional Annotation}} \\ \hline
\begin{tabular}[c]{@{}c@{}}FCCR\\ \cite{csulea2017exploring}\end{tabular} &
  French &
  France &
  126,865 &
  0 &
  - &
  \begin{tabular}[c]{@{}l@{}}Court Decision\\ (6 and 8 w.r.t. two setups)\end{tabular} &
  \begin{tabular}[c]{@{}l@{}}date of the court ruling\\ and law area\end{tabular} \\ \hline
\begin{tabular}[c]{@{}c@{}}CAIL\\ \cite{xiao2018cail2018}\end{tabular} &
  Chinese &
  China &
  2,676,075 &
  0 &
  - &
  \begin{tabular}[c]{@{}l@{}}Law Article (183)\\ Charge (202)\\ Prison Term (integer value)\end{tabular} &
  \begin{tabular}[c]{@{}l@{}}the defendant and\\ penalty of money\end{tabular} \\ \hline
\begin{tabular}[c]{@{}c@{}}ECHR\\ \cite{chalkidis2019neural}\end{tabular} &
  English &
  Europe &
  11,478 &
  0 &
  2421 &
  \begin{tabular}[c]{@{}l@{}}Violation (2) \\ Law Article (66)\end{tabular} &
  case importance \\ \hline
\begin{tabular}[c]{@{}c@{}}ECHR\\ \cite{chalkidis-etal-2021-paragraph}\end{tabular} &
  English &
  Europe &
  11,000 &
  \begin{tabular}[c]{@{}c@{}}50 \\ (fact paragraphs)\end{tabular} &
  - &
  \begin{tabular}[c]{@{}l@{}}Alleged Law Article (40)\\ Violation (2)\\ Law Article (40)\end{tabular} &
  paragraph-level rationale \\ \hline
\begin{tabular}[c]{@{}c@{}}SJP\\ \cite{niklaus2021swiss}\end{tabular} &
  \begin{tabular}[c]{@{}c@{}}German\\ French\\ Italian\end{tabular} &
  Switzerland &
  \begin{tabular}[c]{@{}c@{}}49,883 (German)\\ 31,094 (French)\\ 4,292 (Italian)\end{tabular} &
  \begin{tabular}[c]{@{}c@{}}200 (German) \\ (Court Decision)\end{tabular} &
  850 &
  Court Decision (2) &
  \begin{tabular}[c]{@{}l@{}}publication year\\ legal area\\ canton of origin\end{tabular} \\ \hline
\begin{tabular}[c]{@{}c@{}}ILDC\\ \cite{malik-etal-2021-ildc}\end{tabular} &
  English &
  India &
  34,816 &
  \begin{tabular}[c]{@{}c@{}}56 (Court Decision \\ and Explanation)\end{tabular} &
  3231 &
  Court Decision (2) &
  \begin{tabular}[c]{@{}c@{}}sentence-level explanation\end{tabular} \\ \hline
\begin{tabular}[c]{@{}c@{}}HLDC\\ \cite{kapoor-etal-2022-hldc}\end{tabular} &
  Hindi &
  India &
  340,280 &
  0 &
  764 &
  Bail Prediction (2) &
  extractive summarization \\ \hline
\begin{tabular}[c]{@{}c@{}}BCD\\ \cite{lage2022predicting}\end{tabular} &
  Portuguese &
  Brazil &
  4,043 &
  0 &
  119 &
  \begin{tabular}[c]{@{}l@{}}Court Decision (3)\\ decision’s unanimity status\end{tabular} &
  unanimity label \\ \hline
\begin{tabular}[c]{@{}c@{}}\textbf{(Our dataset)} \\ \textbf{PredEx 2024}\end{tabular} &
  English &
  India &
  15,222 &
  15,222 &
  4,504 &
  \begin{tabular}[c]{@{}l@{}}Court Decision (2)\\ Explanation for Decision\end{tabular} &
  \begin{tabular}[c]{@{}l@{}}expert ratings of generated \\responses for 50 PredEx \\and 54 ILDC experts\end{tabular}  
  
   \\ \hline
\end{tabular}%
}
\caption{Comparative Overview of Widely Used Legal Judgment Prediction Datasets. Entries marked with `-' denote unknown or unavailable information.}
\label{tab:old-data-stat}
\end{table*}

In the evolving landscape of legal technology, the integration of Artificial Intelligence (AI) into the judicial system has emerged as a frontier of immense potential and challenge. The Indian judiciary, characterized by a significant backlog of cases\footnote{\url{https://www.nytimes.com/2024/01/13/world/asia/india-judicial-backlog.html}}, stands to benefit substantially from advancements in AI-assisted legal decision-making. This paper introduces a novel approach to facilitating the legal decision-making process, specifically focusing on the Indian context, in conjunction with explanations for the same. Our work builds upon two foundational studies: \cite{malik-etal-2021-ildc} and \cite{vats-etal-2023-llms}. Our objective is to develop an advanced system capable of predicting judicial outcomes and providing cogent explanations for these predictions. This system leverages a newly compiled dataset, \texttt{PredEx}, of approximately 15,000 annotated legal documents, considerably larger than those used in previous research, particularly in terms of its volume and depth of annotations. Table~\ref{tab:old-data-stat} compares \texttt{PredEx} with other popularly used corpora for legal judgment prediction, highlighting the uniqueness of our dataset in terms of its size and focus on providing explanations. Unlike previous works that predominantly focused on predicting legal outcomes, \texttt{PredEx} introduces the largest annotated dataset for judgment prediction and explanation in the Indian legal context, addressing a critical gap in legal AI research. This dataset enables us to train and refine sophisticated machine learning models, particularly focusing on instruction tuning, to achieve unprecedented accuracy and relevancy in legal judgment prediction.

Our work is distinguished by several key contributions that mark significant advancements in the field of legal AI:
\begin{enumerate}
    \item Publication of a New Annotated Dataset (\texttt{PredEx}): We introduce the largest annotated dataset to date for judgment prediction and explanation in the Indian legal context. This dataset surpasses previous efforts in both scope and depth, providing a more robust foundation for training AI models in legal judgment prediction.
    
    \item Exploration of Instruction-Tuning on Large Language Models (LLMs): Our work goes beyond the traditional methods of fine-tuning conventional transformers. We delve into instruction tuning on LLMs, an approach not extensively explored in previous research, to enhance prediction accuracy.
    
    \item Expert Evaluation and Validation: We employ a rigorous evaluation process, utilizing a Likert score scale to assess the efficacy of our system. This evaluation, conducted on a sample of 50 documents, provides critical insights into the performance of our AI models compared to human expert standards.
\end{enumerate}


Our research aims to provide a comprehensive and sophisticated AI-based system for legal judgment prediction and explanation, specifically tailored for the Indian judiciary. This system is not only a technological advancement but also a step towards addressing the pressing challenge of case backlog in India. We believe our contributions will not only enhance the efficiency and transparency of the legal process but also pave the way for further research and development in AI-assisted legal technology. For the sake of reproducibility, we have made the \texttt{PredEx} dataset and the code for our prediction and explanation models accessible via a GitHub link\footnote{\url{https://github.com/ShubhamKumarNigam/PredEx}}. Additionally, for convenience, we have uploaded the data\footnote{\href{https://huggingface.co/collections/L-NLProc/predex-models-66509d3f4de624770d690a48}{huggingface.co/collections/L-NLProc/predex-models}} and models\footnote{\href{https://huggingface.co/collections/L-NLProc/predex-datasets-6650a75907cc2255eab18d01}{huggingface.co/collections/L-NLProc/predex-datasets}} to Huggingface.

\section{Related Work}
\begin{figure*}[t]
    \centering
    \includegraphics[width=\linewidth]{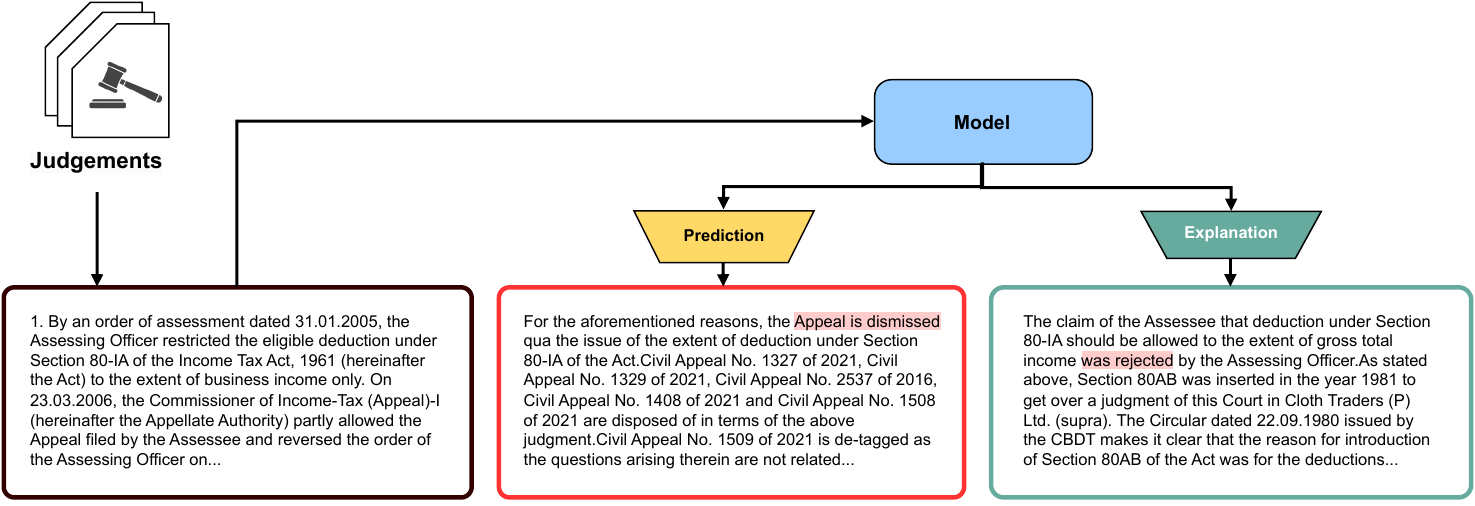}
    \caption{Illustration of the CJPE Task Framework.}
    \label{fig:task-framework}
\end{figure*}
The field of Legal Natural Language Processing (NLP) has witnessed significant advancements, with researchers exploring a variety of complex tasks within the legal domain. A prominent area of focus has been Legal Judgment Prediction (LJP), where the goal is to predict the outcomes of legal cases based on their facts and contexts. Seminal works in this area include the contributions of \cite{zhong2020iteratively}, \cite{malik-etal-2021-ildc}, \cite{aletras2016predicting}, \cite{chen-etal-2019-charge} \cite{long2019automatic}, \cite{xu-etal-2020-distinguish} \cite{10.5555/3367471.3367609}, and \cite{chalkidis2019neural}. These studies have laid the groundwork for understanding the nuances involved in automating legal decision-making processes.

Another key area of research has been the application of Large Language Models (LLMs) in the legal field. The versatility of models such as GPT, BLOOM, FLAN-T5, and LlaMA has been demonstrated in various studies, including those by \cite{vats-etal-2023-llms} \cite{StatutoryReasoningGPT2023} and \cite{barExamGPT4}, highlighting their potential in tasks ranging from statutory reasoning to judgment prediction. However, challenges remain in terms of the acceptability and reliability of LLMs in high-stakes legal contexts. The LegalEval \cite{modi-etal-2023-semeval} workshop further exemplifies the diversity and complexity of legal NLP research, especially on legal judgment prediction and explanation.

Our research utilizes advanced Large Language Models and a comprehensive dataset to create a system that predicts and explains judicial outcomes, enhancing legal text processing and transparency. This work supports legal practitioners and the public, especially in complex systems like India's, and sets the stage for future AI advancements in legal technology.

\section{Task Description}
Our research project aims to advance the Court Judgment Prediction and Explanation (CJPE) task, incorporating insights and methodologies from both \cite{malik-etal-2021-ildc} and \cite{vats-etal-2023-llms}. The CJPE task involves two key sub-tasks: Prediction and Explanation. These tasks are performed sequentially, addressing the critical need to predict legal judgments and provide explanations for these predictions. To provide a visual representation of our task framework, Figure \ref{fig:task-framework} illustrates the overall process of Court Judgment Prediction and Explanation (CJPE) as employed in our study. This figure includes the sequential steps of prediction and explanation. To demonstrate an Indian case structure in Table \ref{case-example} and how explanations are derived from judicial judgments, refer to Figure \ref{fig:annotated-example} in the Appendix. For context, you can view the original case text on which this annotation is based.\footnote{\url{https://indiankanoon.org/doc/97694707/}} This example showcases the detailed process by which our annotators have identified and extracted these critical parts, reflecting the essence of judicial reasoning in each case. Similarly, an Example of an Indian Case Structure Table \ref{case-example}.

\textbf{Prediction Task:} The core of the CJPE task is to predict the outcome of a legal case based on the case proceedings. Given a document \(D\) that includes the case proceedings from the Supreme Court of India (SCI), the task is to predict the decision \(y \in \{0, 1\}\), where `1' signifies the acceptance of the appeal or petition by the appellant or petitioner, and `0' indicates its rejection.

\textbf{Explanation Task:} The second part of the CJPE task involves explaining the predicted decision. Our approach is two-fold, integrating methodologies from both referenced papers:

1. \textbf{Identifying Key Sentences (ILDC for CJPE approach):} Similar to the approach in \cite{malik-etal-2021-ildc}, we focus on identifying and highlighting key sentences or segments within the case proceedings that significantly contributed to the predicted outcome. This method relies on extracting specific parts of the text that are directly related to the decision, providing an evidence-based explanation.

2. \textbf{Generating Abstract Reasoning (LLMs approach):} Drawing from the approach in \cite{vats-etal-2023-llms}, we attempt to generate more abstract reasoning for the prediction. This involves providing zero and few-shot examples to the LLMs to guide them in generating explanations that are not just tied to specific text excerpts but also encompass broader reasoning and legal principles.

Additionally, we introduce a novel aspect to this task by training the LLMs specifically for both prediction and explanation. This training is tailored to enable the models to understand and process legal texts more effectively, improving their capability to predict outcomes and generate relevant explanations.

\section{Dataset}
In our research, we introduce ``PredEx'', significantly differentiating itself from existing datasets in Legal Natural Language Processing (L-NLP), particularly in the context of the Indian judiciary. This dataset is designed to address the limitations of previous datasets, which primarily focused on prediction tasks and offered limited annotations for explanations.

\subsection{Dataset Compilation} In the Data Compilation process, we initially gathered a substantial corpus of about 20,000 court judgments randomly from the Supreme Court of India and various High Courts, utilizing the IndianKanoon website\footnote{\url{https://indiankanoon.org/}}, a legal search engine widely recognized for its comprehensive database of Indian legal documents. The corpus underwent a meticulous annotation process, where our team of legal experts focused on annotating explanations for the judgments. These annotations involved identifying and highlighting key sentences or segments within the case proceedings that significantly influenced the predicted outcomes, as well as providing reasoning for the judgments. Through this process, the original corpus was distilled to approximately 16,000 case files, each richly annotated with expert legal explanations.

Our approach in compiling the PredEx dataset was to randomly select cases, ensuring a broad representation across various types of judgments and legal decisions. This method was deliberately chosen to avoid bias towards any specific temporal aspect or domain. By adopting a random selection process, we aimed to capture the diverse nature of legal cases in the Indian judiciary system. This diversity was crucial to cover various aspects of law and legal decision-making, thereby enhancing the representativeness and applicability of our dataset for training AI models in legal judgment prediction and explanation.

Subsequent to the annotation phase, we undertook a preprocessing step to refine the dataset further. This preprocessing involved the removal of cases that were either too brief or where the final decision segments were challenging to discern. Such preprocessing is crucial for ensuring the quality and consistency of the data, particularly for training robust and reliable AI models; otherwise, it could introduce noise or bias into the model training. As a result of this preprocessing, the total number of case files in our dataset was reduced to 15,222 and is further divided into training and testing sets. We adopted an 80-20 split ratio for this purpose, ensuring a substantial volume of data for model training while still retaining a robust set for testing. Specifically, the training set consists of 12,178 documents, and the test set comprises 3,044 documents. 

In terms of balancing the test set, special attention was given to ensure fairness and representativeness in model evaluation. We carefully curated the test set to include a diverse range of case outcomes, such as different types of judgments and legal decisions. This diversity was not just in terms of the nature of cases but also in terms of the outcomes - for instance, balancing cases where appeals were accepted versus those that were dismissed. Such a balanced composition is crucial in avoiding biases towards any particular type of judgment and ensures that our AI models are tested against a wide spectrum of legal scenarios. This balanced nature of the test set is particularly important for maintaining the validity of our experiments and for ensuring the reliability and generalizability of our model's performance. These carefully processed and curated case files now form the core of our \texttt{PredEx} dataset, offering a rich resource for the Court Judgment Prediction and Explanation (CJPE) task. Detailed statistics of the final dataset, post-preprocessing, are presented in the following Table \ref{data-stats}.

\begin{table}[t]
  \centering
\resizebox{\columnwidth}{!}{%
  \begin{tabular}{lcc}
    \toprule
    & \textbf{Train} & \textbf{Test} \\
    \midrule
    No. of documents & 12,178 & 3,044 \\
    Average no. of tokens & 4,586 & 4,422 \\
    Minimum no. of tokens & 176 & 184 \\
    Maximum no. of tokens & 117,733 & 83,657 \\
    Acceptance percentage & 53.44\% & 50.00\% \\
    \bottomrule
  \end{tabular}
  }
    \caption{PredEx Statistics.}
  \label{data-stats}
\end{table}

\subsection{Annotation Process}
\subsubsection{Expert Involvement} We engaged a team of 10 legal experts, primarily law students in their 3rd and 4th years, from various Indian law colleges. These experts were selected based on their academic standing and understanding of legal processes, ensuring high-quality annotations.

\subsubsection{Annotation Timeline} The annotation process spanned from April 1, 2022, to October 30, 2023. This extensive period allowed for meticulous and thorough annotation, considering the complexity and detail required in legal document analysis.

\subsubsection{Work Allocation}
In our annotation process, each student was assigned around 30 judgment documents weekly, striking a balance between efficiency and the need for thorough, accurate annotations. This workload allocation enabled students to devote adequate time to each document, fostering precise and insightful annotations. 
\subsubsection{Role of Student Annotators}
The role of our student annotators was to meticulously identify and extract specific segments from the judgments that were pivotal to the judge's reasoning, rather than interpreting or analyzing these segments with their own legal reasoning. Their task was to pinpoint these key sections accurately, ensuring that the extracts faithfully represented the judicial reasoning as stated in the case documents. This extractive approach was critical to maintain the integrity and authenticity of the annotations, allowing the dataset to accurately reflect the content of the original legal texts without the introduction of subjective interpretations by the annotators.

%
\subsubsection{Annotation Quality Control Mechanism}
To guarantee the accuracy and consistency of the annotations, we implemented a comprehensive quality control system. Initially, each document was reviewed by a single annotator to ensure a consistent interpretation of the judicial content. Recognizing the complexity of legal texts and the potential for subjective interpretation, we established several layers of review to enhance the reliability of our annotations:
\begin{itemize}
    \item \textbf{Senior Expert Review:} Any disagreements or uncertainties in annotations were promptly escalated to a specialized review panel led by senior legal experts. These experts not only provided additional scrutiny but also mediated discrepancies among the initial annotations. Their extensive experience in legal practice and education enabled them to provide decisive and informed resolutions to any contentious or ambiguous annotations.
    \item \textbf{Regular Training and Meetings:} To further ensure consistency across annotations, regular training sessions and review meetings were conducted. These sessions served to align annotators on the legal framework and annotation criteria, reducing variability and enhancing the uniformity of the annotation process. Training included detailed discussions on identifying key legal arguments and the rationale within the judgments, which are critical for both the prediction and explanation aspects of our dataset.
\end{itemize}

This rigorous quality control process has ensured that our dataset meets high standards of reliability and validity. The annotations not only reflect the factual content of the legal decisions but also the detailed reasoning behind these judgments, making our dataset a robust resource for training and evaluating AI models in legal judgment prediction and explanation.

\subsubsection{Focus on Prediction and Explanations}
Diverging from previous datasets that primarily concentrate on the task of prediction, our \texttt{PredEx} dataset spans both prediction and explanations. The annotations in our dataset serve a dual purpose. Firstly, they identify the outcomes of the cases, fulfilling the prediction aspect. More importantly, they go a step further by providing detailed explanations behind these outcomes. These explanations elucidate the rationale or the legal reasoning that underpins the judgments. This dual emphasis on prediction and explanations fills a significant void in existing legal datasets. Typically, in other datasets, the aspect of explanation is either absent or not explored in depth. By contrast, \texttt{PredEx} enriches the field of legal AI with comprehensive annotations that shed light not just on what the judicial decisions are, but crucially, why these decisions were made. This focus on explanations is particularly vital, as it contributes to a more transparent and understandable AI-driven legal decision-making process.

\subsubsection{Largest Explainable Dataset} As a result of this extensive and detailed annotation process, we are releasing what is arguably the largest annotated dataset for legal judgment prediction and explanation in the Indian context. The size and comprehensiveness of this dataset set it apart from existing datasets in the field.

Our dataset represents a significant advancement in legal NLP, particularly for research and applications pertaining to the Indian judiciary. By providing a large-scale, richly annotated dataset that encompasses both prediction and explanation, we aim to facilitate more nuanced and sophisticated AI models capable of understanding and interpreting legal texts in a manner akin to human legal experts. This dataset is not only a resource for advancing AI technology in the legal domain but also a step towards enhancing transparency and accountability in AI-assisted legal decision-making.

\section{Methodology}
This section outlines the methodology employed in our research for the tasks of Judgment Prediction and Judgment Prediction with Explanation.

\subsection{Judgment Prediction}
\subsubsection{Language Model based}
In our approach, we utilized several language models including InCaseLaw, InLegalBERT \cite{paul-2022-pretraining}, XLNet (large) \cite{yang2019xlnet}, and Roberta (large) \cite{liu2019roberta} as baselines for binary classification. Due to the length constraints of complete judgments, which exceed the token capacity of these models, we adopted a chunking strategy. Each document was divided into 512-token chunks using a moving window approach with a 100-token overlap to preserve textual context. For model training, we used a batch size of 16, the Adam optimizer \cite{kingma2014adam}, and a learning rate of 2e-6. The training was conducted over 5 epochs on the PredEx train dataset. The remaining hyperparameters were set to their default values as provided by the HuggingFace library.

\subsubsection{Large Language Model based}
For utilizing LLMs in prediction, we employed two strategies: one involving prediction instructions only, and the other combining prediction with explanation instructions. Various models like Zephyr \cite{tunstall2023zephyr}, Gemini 1.0 Pro \cite{team2023gemini}, Llama-2-7B \cite{touvron2023llama}, and Llama-2-7B with instruction-tuning were used. We followed the prompts and instruction-tuning approaches published by \cite{vats-etal-2023-llms} in a few-shot setup, and used the PredEx training data for instruction-tuning.

\subsubsection{Prompts Used}
For inference, we utilized prompts published by \cite{vats-etal-2023-llms} and employed Template 2 in a zero and few-shot setup exclusively for prediction tasks, as detailed in Table \ref{tab:judgment_prediction_prompts_few} in the Appendix. These prompts provide a case description alongside a gold standard prediction label, directing the LLM to generate the judicial decision. For instruction tuning, we employed our custom prompts for prediction tasks, which are listed in Table \ref{tab:judgment_prediction_prompts_zero} in the Appendix of this paper.

\subsubsection{Instruction-Set}
We developed 16 instruction sets using ChatGPT4 (DALL-E), validated by legal experts and then used for PredEx training data for instruction tuning. Given the token limit of 4096 in LLMs, we selected the last 1000 words from each document to fit within this constraint. This choice is supported by findings from \cite{malik-etal-2021-ildc} who achieved optimal results using the last 512 tokens of judgments. The input comprised the case proceedings and case decision and a random selection of instructions, with the output being the case outcome prediction. For a comprehensive understanding of our methodology and the full range of instructions used, we have included the complete list of all 16 instruction sets in Table~\ref{Instruction-sets} located in the Appendix of this paper.

\subsection{Judgment Prediction with Explanation}
For this task, we employed the same LLMs with settings similar to the Judgment Prediction task, but with different instructions focusing on both prediction and explanation.

\subsubsection{Prompts used}
In our approach, we also adopted prompts from \cite{vats-etal-2023-llms}, specifically utilizing Template 1 for the combined task of judgment prediction and explanation. This template, detailed in Table \ref{tab:judgment_prediction_prompts_few} in the Appendix, expands on the prediction-only format of Template 2 by including an explanation component. Here also, for instruction tuning, we employed our custom prompts for prediction with explanation tasks, which are listed in Table \ref{tab:judgment_prediction_prompts_zero} in the Appendix of this paper. In this setup, the LLM is instructed not only to predict the outcome of a case but also to articulate the reasoning behind the decision. The precise formatting and examples of how predictions and explanations are structured and solicited from the LLM can be viewed in the referenced table.

\subsubsection{Instruction-Set}
For judgment prediction with explanation, we created 16 instruction sets using ChatGPT4 (DALL-E), also validated by legal experts. This time, the input included case proceedings, decisions, and reasoning, with randomly chosen instructions, and the output being the case outcome prediction with explanation. For a comprehensive view of all 16 instruction sets, we have included the full list in Table~\ref{Instruction-sets} in the Appendix of this paper.

\section{Evaluation Metrics}
\label{sec:performance_metrics}
In our study, We report Macro Precision, Macro Racall, Macro F1, and Accuracy on the PredEx judgment prediction test dataset and employ a multifaceted approach to evaluate the performance of our models on the PredEx judgment explanation test dataset. Our evaluation metrics encompass both quantitative and qualitative methods, ensuring a thorough assessment of the model's capabilities in both prediction and explanation tasks.

\begin{enumerate}
    \item \textbf{Lexical Based Evaluation:}
    We utilized lexical similarity metrics such as Rouge scores (Rouge-1, Rouge-2, and Rouge-L) ~\cite{lin-2004-rouge}, BLEU ~\cite{papineni-etal-2002-bleu}, and METEOR ~\cite{banerjee-lavie-2005-meteor}. These metrics assess the similarity between the generated explanations and the reference texts based on word overlap and order, providing an insight into the lexical accuracy of the model outputs.

    \item \textbf{Semantic Similarity Based Method:}
    To capture the semantic essence of the generated explanation, we employed BERTScore ~\cite{BERTScore}, which measures the semantic similarity between the generated and ground truth explanations. Additionally, we used BLANC ~\cite{blanc} to estimate the quality of generated explanations in the absence of a gold standard, offering a perspective on the model's ability to generate semantically rich and contextually relevant text.

    \item \textbf{Expert Evaluation:}
    Human evaluation played a crucial role in our assessment. Legal experts reviewed the explanations generated by the models and rated them on a 1–5 Likert scale based on their accuracy, relevance, and completeness. The criteria for the rating scale were as follows:
    \begin{enumerate}[label=\arabic*.]
        \item The explanation is entirely incorrect or fails to provide any relevant information.
        \item The model's response is irrelevant or shows misunderstanding of the case judgment.
        \item The explanation is partially accurate but misses critical details.
        \item The response is comparable and relevant to the ground truth.
        \item The explanation is completely accurate, relevant, and potentially superior to the expert's explanation.
    \end{enumerate}
    
\end{enumerate}

\section{Results and Analysis}

\subsection{Judgment Prediction}
Our experiments, as detailed in Table \ref{prediction-table}, reveal interesting insights into the performance of various models on the PredEx test data. Notably, Roberta emerges as the top performer, outstripping even the Large Language Models (LLMs). This suggests that traditional language models might be more adept at analyzing and predicting outcomes in legal documents compared to generative-based models. Even among the generative models, the few-shot Llama-2-7B model surpassed the fine-tuned Zephyr model, which is surprising given Zephyr's supervised fine-tuning (SFT) approach and its reinforcement learning training on general corpora. It appears that the Llama-2-7B models, both instruction-tuned for prediction and prediction with explanation tasks, show promising results in this domain.


\begin{table}[t]
\centering
\resizebox{\columnwidth}{!}{%
\begin{tabular}{|
>{}c |lc
>{}c c
>{}c |}
\hline
\multicolumn{1}{|l|}{} &
  \multicolumn{1}{l|}{\textbf{Models}} &
  \multicolumn{1}{c|}{\textbf{\begin{tabular}[c]{@{}c@{}}Macro \\ Precision\end{tabular}}} &
  \multicolumn{1}{c|}{\textbf{\begin{tabular}[c]{@{}c@{}}Macro \\ Recall\end{tabular}}} &
  \multicolumn{1}{c|}{\textbf{\begin{tabular}[c]{@{}c@{}}Macro \\ F1\end{tabular}}} &
  \textbf{Accuracy} \\ \hline
\multicolumn{1}{|l|}{}& 
  \multicolumn{5}{c|}{\textbf{Prediction only}} \\ \hline
 &
  \multicolumn{1}{l|}{InLegalBert} &
  \multicolumn{1}{c|}{0.7546} &
  \multicolumn{1}{c|}{0.7526} &
  \multicolumn{1}{c|}{0.7536} &
  0.7526 \\ \cline{2-6} 
 &
  \multicolumn{1}{l|}{InCaseLaw} &
  \multicolumn{1}{c|}{0.7421} &
  \multicolumn{1}{c|}{0.7395} &
  \multicolumn{1}{c|}{0.7408} &
  0.7395 \\ \cline{2-6} 
 &
  \multicolumn{1}{l|}{XLNet Large} &
  \multicolumn{1}{c|}{0.7736} &
  \multicolumn{1}{c|}{0.7707} &
  \multicolumn{1}{c|}{0.7722} &
  0.7707 \\ \cline{2-6} 
\multirow{-4}{*}{\textbf{\begin{tabular}[c]{@{}c@{}}LM\\ Based\end{tabular}}} &
  \multicolumn{1}{l|}{RoBerta Large} &
  \multicolumn{1}{c|}{\textbf{0.7831}} &
  \multicolumn{1}{c|}{\textbf{0.7822}} &
  \multicolumn{1}{c|}{\textbf{0.7827}} &
  \textbf{0.7822} \\ \hline
 &
  \multicolumn{1}{l|}{Zephyr} &
  \multicolumn{1}{c|}{{ 0.5347}} &
  \multicolumn{1}{c|}{{ 0.5295}} &
  \multicolumn{1}{c|}{{ 0.5119}} &
  { 0.5309} \\ \cline{2-6} 
 &
  \multicolumn{1}{l|}{Gemini pro} &
  \multicolumn{1}{c|}{{ \textbf{0.5976}}} &
  \multicolumn{1}{c|}{{ \textbf{0.5803}}} &
  \multicolumn{1}{c|}{{ 0.5610}} &
  { \textbf{0.5808}} \\ \cline{2-6} 
 &
  \multicolumn{1}{l|}{Llama-2-7B} &
  \multicolumn{1}{c|}{{ 0.5732}} &
  \multicolumn{1}{c|}{{ 0.5723}} &
  \multicolumn{1}{c|}{{ \textbf{0.5713}}} &
  { 0.5726} \\ \cline{2-6} 
 &
  \multicolumn{1}{l|}{\begin{tabular}[c]{@{}l@{}}Llama-2-7B\\ Instruction-tuning \\ on prediction task\end{tabular}} &
  \multicolumn{1}{c|}{{ 0.5186}} &
  \multicolumn{1}{c|}{{ 0.5177}} &
  \multicolumn{1}{c|}{{ 0.5117}} &
  { 0.5177} \\ \cline{2-6} 
\multirow{-5}{*}{\textbf{\begin{tabular}[c]{@{}c@{}}LLM \\ Based\end{tabular}}} &
  \multicolumn{1}{l|}{\begin{tabular}[c]{@{}l@{}}Llama-2-7B\\ Instruction-tuning \\ on prediction with \\ explanation task\end{tabular}} &
  \multicolumn{1}{c|}{{ 0.5195}} &
  \multicolumn{1}{c|}{{ 0.5185}} &
  \multicolumn{1}{c|}{{ 0.5127}} &
  { 0.5190} \\ \hline
\multicolumn{1}{|l|}{} &
  \multicolumn{5}{c|}{\textbf{Prediction with explanation on PredEx}} \\ \hline
 &
  \multicolumn{1}{l|}{Gemini pro} &
  \multicolumn{1}{c|}{{ 0.5184}} &
  \multicolumn{1}{c|}{{ 0.5154}} &
  \multicolumn{1}{c|}{{ 0.4908}} &
  { 0.5081} \\ \cline{2-6} 
 &
  \multicolumn{1}{l|}{Llama-2-7B} &
  \multicolumn{1}{c|}{{ 0.5087}} &
  \multicolumn{1}{c|}{{ 0.5017}} &
  \multicolumn{1}{c|}{0.3772} &
  { 0.5025} \\ \cline{2-6} 
\multirow{-3}{*}{\textbf{\begin{tabular}[c]{@{}c@{}}LLM \\ Based\end{tabular}}} &
  \multicolumn{1}{l|}{\begin{tabular}[c]{@{}l@{}}Llama-2-7B\\ Instruction-tuning \\ on prediction with \\ explanation task\end{tabular}} &
  \multicolumn{1}{c|}{{ \textbf{0.5254}}} &
  \multicolumn{1}{c|}{{ \textbf{0.5215}}} &
  \multicolumn{1}{c|}{{ \textbf{0.5031}}} &
  { \textbf{0.5224}} \\ \hline
\multicolumn{1}{|l|}{} &
  \multicolumn{5}{c|}{\textbf{Prediction with explanation on ILDC expert}} \\ \hline
 &
  \multicolumn{1}{l|}{Llama-2-7B} &
  \multicolumn{1}{c|}{{ 0.3125}} &
  \multicolumn{1}{c|}{{ 0.4259}} &
  \multicolumn{1}{c|}{{ 0.3236}} &
  { 0.4259} \\ \cline{2-6} 
\multirow{-2}{*}{\textbf{\begin{tabular}[c]{@{}c@{}}LLM \\ Based\end{tabular}}} &
  \multicolumn{1}{l|}{\begin{tabular}[c]{@{}l@{}}Llama-2-7B\\ Instruction-tuning \\ on prediction with \\ explanation task\end{tabular}} &
  \multicolumn{1}{c|}{{ \textbf{0.5750}}} &
  \multicolumn{1}{c|}{{ \textbf{0.5741}}} &
  \multicolumn{1}{c|}{{ \textbf{0.5728}}} &
  { \textbf{0.5741}} \\ \hline
\end{tabular}%
}
\caption{Judgement prediction results. The best results are shown in bold.}
\label{prediction-table}
\end{table}


\begin{table*}[t]
\centering
\resizebox{\linewidth}{!}{%
\begin{tabular}{|l|
>{}c 
>{}c 
>{}c 
>{}c 
>{}c 
>{}c cc|}
\hline
 &
  \multicolumn{5}{c|}{\textbf{Lexical Based Evaluation}} &
  \multicolumn{2}{c|}{\textbf{Semantic Evaluation}} &
  \textbf{Expert Evaluation} \\ \cline{2-9} 
\multirow{-2}{*}{\textbf{Models}} &
  \multicolumn{1}{c|}{\textbf{Rouge-1}} &
  \multicolumn{1}{c|}{\textbf{Rouge-2}} &
  \multicolumn{1}{c|}{\textbf{Rouge-L}} &
  \multicolumn{1}{c|}{\textbf{BLEU}} &
  \multicolumn{1}{c|}{\textbf{METEOR}} &
  \multicolumn{1}{c|}{\textbf{BERTScore}} &
  \multicolumn{1}{c|}{\textbf{BLANC}} &
  \textbf{Rating Score} \\ \hline
 &
  \multicolumn{8}{c|}{\textbf{Prediction with explanation on PredEx}} \\ \hline
Gemini pro &
  \multicolumn{1}{c|}{{ 0.3099}} &
  \multicolumn{1}{c|}{{ 0.2428}} &
  \multicolumn{1}{c|}{{ 0.2593}} &
  \multicolumn{1}{c|}{{ 0.0826}} &
  \multicolumn{1}{c|}{{ 0.1870}} &
  \multicolumn{1}{c|}{{ 0.6329}} &
  \multicolumn{1}{c|}{{ 0.1715}} &
  2.24 \\ \hline
Llama-2-7B &
  \multicolumn{1}{c|}{{ 0.3211}} &
  \multicolumn{1}{c|}{{ 0.1886}} &
  \multicolumn{1}{c|}{{ 0.2109}} &
  \multicolumn{1}{c|}{{ 0.0599}} &
  \multicolumn{1}{c|}{{ 0.1760}} &
  \multicolumn{1}{c|}{{ 0.6191}} &
  \multicolumn{1}{c|}{{ 0.1507}} &
  { \textbf{3.06}} \\ \hline
\begin{tabular}[c]{@{}l@{}}Llama-2-7B\\ Instruction-tuning \\ on prediction with \\ explanation task\end{tabular} &
  \multicolumn{1}{c|}{{ \textbf{0.4972}}} &
  \multicolumn{1}{c|}{{ \textbf{0.4321}}} &
  \multicolumn{1}{c|}{{ \textbf{0.4399}}} &
  \multicolumn{1}{c|}{{ \textbf{0.2531}}} &
  \multicolumn{1}{c|}{{ \textbf{0.3630}}} &
  \multicolumn{1}{c|}{{ \textbf{0.6909}}} &
  \multicolumn{1}{c|}{{ \textbf{0.2844}}} &
  { 2.84} \\ \hline
 &
  \multicolumn{8}{c|}{\textbf{Prediction with explanation on ILDC expert \cite{vats-etal-2023-llms, malik-etal-2021-ildc}}} \\ \hline
\begin{tabular}[c]{@{}l@{}}GPT 3.5 turbo\\ (Reproduced)\end{tabular} &
  \multicolumn{1}{c|}{{ \textbf{0.5383}}} &
  \multicolumn{1}{c|}{{ \textbf{0.4267}}} &
  \multicolumn{1}{c|}{{ \textbf{0.4541}}} &
  \multicolumn{1}{c|}{{ 0.2842}} &
  \multicolumn{1}{c|}{{ 0.4685}} &
  \multicolumn{1}{c|}{{ \textbf{0.7273}}} &
  \multicolumn{1}{c|}{0.3394} &
  3.6$^*$ \\ \hline
Llama-2-7B &
  \multicolumn{1}{c|}{{ 0.4526}} &
  \multicolumn{1}{c|}{{ 0.2454}} &
  \multicolumn{1}{c|}{{ 0.2957}} &
  \multicolumn{1}{c|}{{ 0.1485}} &
  \multicolumn{1}{c|}{{ 0.3440}} &
  \multicolumn{1}{c|}{{ 0.6464}} &
  \multicolumn{1}{c|}{{ 0.2212}} &
  { \textbf{3.65}} \\ \hline
\begin{tabular}[c]{@{}l@{}}Llama-2-7B\\ Instruction-tuning \\ on prediction with \\ explanation task\end{tabular} &
  \multicolumn{1}{c|}{{ 0.4939}} &
  \multicolumn{1}{c|}{{ 0.3805}} &
  \multicolumn{1}{c|}{{ 0.3969}} &
  \multicolumn{1}{c|}{{ \textbf{0.2918}}} &
  \multicolumn{1}{c|}{{ \textbf{0.5075}}} &
  \multicolumn{1}{c|}{{ 0.6891}} &
  \multicolumn{1}{c|}{{ \textbf{0.3636}}} &
  { 3.30} \\ \hline
\end{tabular}%
}
\caption{Explanation performance comparison of various model combinations for judgment prediction across different evaluation metrics, with the highest score in each metric in bold. Entries marked with $*$ denote normalized value.}
\label{tab:explanation-table}
\end{table*}


\subsection{Judgment Prediction with Explanation}
The results, as shown in Table \ref{tab:explanation-table}, provide valuable insights into the performance of machine-generated explanations versus expert explanations across a range of models. These assessments include lexical-based, semantic, and expert evaluations on the PredEx test data. To augment our evaluation process, we also incorporated a comparison with the instruction-tuned models on the 54 ILDC\_expert \cite{malik-etal-2021-ildc} dataset. This dataset, to our knowledge, represents the largest collection of legal expert-annotated data available for Indian cases, offering a valuable benchmark for assessing the performance of our models. This multi-faceted evaluation offers a comprehensive understanding of the models' capabilities in generating explanations.

Given the expense and time required to obtain legal expert annotations, we carefully sampled 50 cases from our dataset for Likert score evaluations by legal experts. This sampling strategy was chosen to provide a representative and manageable subset of cases for in-depth expert analysis, while also considering the practical constraints associated with expert-driven evaluations.

\subsection{Lexical Based Evaluation}
In the lexical-based evaluation, the performance of LLMs in generating explanations shows that verbatim matches are not at a satisfactory level. However, it's important to note that these metrics, while valuable, do not fully encapsulate the models' proficiency in analyzing cases, predicting outcomes, and generating reasoning. Thus, we turn to Semantic Similarity-Based Evaluation and Expert Score Evaluation for a more thorough assessment.

\subsection{Semantic Evaluation}
Semantic evaluation, particularly the BERTScore, indicates better alignment of the explanations with the gold standard, suggesting a good semantic understanding in the generated explanations. The Llama-2-7B model with instruction-tuning for prediction and explanation tasks excels in semantic similarity. Nevertheless, lower scores in open-source models point to challenges in accurately generating case analysis, predictions, and reasoning. It's crucial to recognize that generative models may exhibit hallucination issues, not entirely captured by this metric, necessitating manual evaluation by legal experts for a more complete assessment.

\subsection{Expert Evaluation}
\label{subsec:expert_evaluation}
Evaluating generative models in the legal judgment prediction task with explanation requires domain-specific expertise. The expert evaluation, detailed in Table \ref{expert-scores}, shows that the Llama-2-7B model with instruction-tuning performs notably well, although it sometimes produces truncated or repetitive responses. Despite these limitations, the instruction-tuned model demonstrates fewer non-factual responses and better overall explanation quality compared to other pre-trained models. Interestingly, models with well-designed prompts for explanation generation displayed enhanced performance without instances of hallucination.

The expert ratings, as reflected in Table \ref{expert-scores}, further underscore the efficacy of our instruction-tuned model, which even surpasses the quality of explanations provided by legal professionals (achieving a rating score of 4). This underlines the potential of generative models, particularly those leveraging our instruction-tuning approach, in generating accurate and relevant legal explanations. The average expert rating scores, presented in Table \ref{tab:explanation-table}, corroborate the superiority of our generative models over other approaches.


\begin{table}[t]
\centering
\resizebox{\columnwidth}{!}{%
\begin{tabular}{|l|ccccc|}
\hline
 
 &
  \multicolumn{5}{c|}{\textbf{Rating Score}} \\ \hline
 
 &
  \multicolumn{1}{c|}{\textbf{1}} &
  \multicolumn{1}{c|}{\textbf{2}} &
  \multicolumn{1}{c|}{\textbf{3}} &
  \multicolumn{1}{c|}{\textbf{4}} &
  \textbf{5} \\ \cline{2-6} 
 
\multirow{-2}{*}{\textbf{Generative Models}} &
  \multicolumn{5}{c|}{\textbf{PredEx}} \\ \hline
Llama-2-7B &
  \multicolumn{1}{c|}{{ 2}} &
  \multicolumn{1}{c|}{{ 11}} &
  \multicolumn{1}{c|}{{ 22}} &
  \multicolumn{1}{c|}{{ 12}} &
  { 3} \\ \hline
\begin{tabular}[c]{@{}l@{}}Llama-2-7B\\ Instruction-tuned\end{tabular} &
  \multicolumn{1}{c|}{{ 5}} &
  \multicolumn{1}{c|}{13} &
  \multicolumn{1}{c|}{18} &
  \multicolumn{1}{c|}{13} &
  1 \\ \hline
 &
  \multicolumn{5}{c|}{\textbf{ILDC expert}} \\ \hline
Llama-2-7B &
  \multicolumn{1}{c|}{0} &
  \multicolumn{1}{c|}{9} &
  \multicolumn{1}{c|}{22} &
  \multicolumn{1}{c|}{21} &
  2 \\ \hline
\begin{tabular}[c]{@{}l@{}}Llama-2-7B\\ Instruction-tuned\end{tabular} &
  \multicolumn{1}{c|}{2} &
  \multicolumn{1}{c|}{3} &
  \multicolumn{1}{c|}{16} &
  \multicolumn{1}{c|}{24} &
  9 \\ \hline
\end{tabular}%
}
\caption{Distribution of Expert Rating Scores for Generative Models on PredEx and ILDC Expert Data.}
\label{expert-scores}
\end{table}
\subsection{Hallucination}
\label{subsec:hallucination}
We address the issue of hallucinations in model-generated text, a common challenge in using large language models for generating legal judgments. Hallucinations refer to instances where the model generates false or irrelevant information that is not supported by the input data. To combat this, we have implemented a fine-tuning strategy that significantly reduces these errors. A detailed comparative analysis in the Appendix \ref{sec:Hallucination-example} showcases these strategies and their effectiveness. This analysis demonstrates how fine-tuning and instruction-tuning specifically tailored to the legal domain can mitigate hallucinations, providing clearer, more accurate, and legally coherent outputs.

\section{Conclusions and Future Work}
We introduced \texttt{PredEx}, the largest dataset for legal judgment prediction and explanation in this context, marking a significant advancement over previous datasets. Our research explored instruction tuning on Large Language Models (LLMs), showing promise in improving prediction accuracy and explanatory depth.

Looking ahead, our focus will be on training Indian Legal domain-specific Large Language Models. This approach will ensure that the models are ingrained with domain-specific knowledge, crucial for tasks like legal judgment prediction with explanations. Furthermore, we plan to undertake Supervised Fine-Tuning (SFT) on various downstream tasks, including the judgment prediction with explanation task. Another key objective will be to incorporate contextual understanding into the models to mitigate issues like hallucinated responses, a common challenge with generative models.

The question remains as we advance in this field: How ready is the State-of-the-Art to aid in explainable judgment prediction? Our future efforts aim to answer this question by refining the capabilities of AI in legal applications, making a significant contribution to the evolving field of AI-assisted legal judgment prediction and explanation. The ultimate goal is to develop AI tools that can not only alleviate the backlog in the Indian judiciary but also deliver justice efficiently and transparently. To further enhance the accuracy and reliability of our system, we plan to implement a Reinforcement Learning from Human Feedback (RLHF) pipeline. This pipeline aims to refine the model's predictions and explanations based on human feedback, ensuring that the outputs align more closely with expert legal understanding and reasoning. The inclusion of RLHF represents a significant advancement in developing AI systems for legal judgment prediction and explanation, as it allows for continuous improvement and adaptation based on real-world feedback and interactions.

\section*{Limitations}
Our study faced several significant limitations that impacted our approach and findings. A primary constraint was the token limitation and high subscription charges for paid cloud services, which restricted our ability to perform inference and fine-tuning on larger parametric models, particularly those with 70B or 40B parameters. This limitation likely curtailed our exploration of the full capabilities of these advanced models, which could have provided deeper insights or enhanced performance.

Another critical limitation was the resource-intensive nature of obtaining legal expert annotations. Due to the high costs and extensive time required for this process, it was not feasible for us to obtain expert evaluations for the entire PredEx test dataset. Consequently, we opted to sample 50 random documents for expert review and Likert score evaluations. While necessary, this approach potentially limits the breadth and depth of our expert-based evaluation, as it does not encompass the entire dataset.

In terms of the effectiveness of Large Language Models (LLMs) in the legal domain, our findings suggest that while these models are proficient in conversational contexts, their applicability in logic or knowledge-intensive tasks like legal judgment prediction and explanation is less convincing. Analyzing lengthy legal documents and generating predictions with explanations poses a significant challenge for generative-based models. This is particularly true in cases where the models need to process and understand complex legal reasoning and contexts.

Furthermore, the performance of the open-source baseline model, which was intended to jointly predict and generate explanations, did not meet our expectations. This underperformance could be attributed to the token limitations imposed during our study. By only using the last 1000 tokens of documents for fine-tuning, there is a possibility that the model did not fully grasp the entire context of the cases. Moreover, our fine-tuned models frequently produced truncated responses due to the 512-token limit set for generation. This limitation may have hindered the models' ability to generate comprehensive and nuanced explanations.

Lastly, the pre-trained models used in our study inherently lacked detailed knowledge specific to Indian legal cases. Even after undergoing tuning processes, these models struggled to generate explanations that paralleled the depth and specificity of human-like legal reasoning. This shortfall highlights the challenge of adapting general AI models to specialized domains such as law, where domain-specific knowledge and reasoning are crucial.

These limitations underscore the challenges in applying LLMs to complex and specialized tasks like legal judgment prediction and explanation. They also highlight the necessity for continued research and development efforts aimed at enhancing the capabilities of AI models in interpreting and understanding legal documents and contexts.

\section*{Ethics Statement}
Ethical conduct was a cornerstone in our research, especially considering the sensitive nature of the data and the methodologies involved. In collecting and annotating the \texttt{PredEx} dataset, we ensured that the law students involved in the annotation process were treated fairly and compensated appropriately. Their consent was obtained for all participation, and while they made significant contributions to the dataset, they are not listed as authors of this paper. This distinction is made to acknowledge their contribution while also maintaining the academic integrity of the publication process.

Significantly, the senior legal expert who played a pivotal role in mentoring the annotation process, as well as providing guidance on the Likert rating system and evaluating the generated explanations for both the PredEx and ILDC datasets, is credited as one of the authors of this paper. This inclusion reflects the expert's substantial intellectual contribution to the research, in line with ethical norms and authorship guidelines in academic publishing.

Moreover, for the computational resources used in this study, we adhered to ethical standards by duly paying the subscription fees for Google Colab Pro. This payment ensured legitimate access to the necessary paid cloud services, which were instrumental in the development and testing of our AI models. We believe in supporting the services and platforms that enable research like ours, and this includes the responsible financial support of technology providers.

In summary, our approach to ethics encompassed not only the respectful and fair treatment of all individuals involved but also the adherence to legal and financial obligations. This comprehensive ethical stance underscores our commitment to conducting research that is not only innovative and impactful but also responsible and respectful of all parties involved.

\newpage
\bibliography{anthology,custom}

\begin{thebibliography}{30}
\expandafter\ifx\csname natexlab\endcsname\relax\def\natexlab#1{#1}\fi

\bibitem[{Aletras et~al.(2016)Aletras, Tsarapatsanis, Preo{\c{t}}iuc-Pietro, and Lampos}]{aletras2016predicting}
Nikolaos Aletras, Dimitrios Tsarapatsanis, Daniel Preo{\c{t}}iuc-Pietro, and Vasileios Lampos. 2016.
\newblock Predicting judicial decisions of the european court of human rights: A natural language processing perspective.
\newblock \emph{PeerJ computer science}, 2:e93.

\bibitem[{Banerjee and Lavie(2005)}]{banerjee-lavie-2005-meteor}
Satanjeev Banerjee and Alon Lavie. 2005.
\newblock \href {https://aclanthology.org/W05-0909} {{METEOR}: An automatic metric for {MT} evaluation with improved correlation with human judgments}.
\newblock In \emph{Proceedings of the {ACL} Workshop on Intrinsic and Extrinsic Evaluation Measures for Machine Translation and/or Summarization}, pages 65--72, Ann Arbor, Michigan. Association for Computational Linguistics.

\bibitem[{Blair-Stanek et~al.(2023)Blair-Stanek, Holzenberger, and Durme}]{StatutoryReasoningGPT2023}
Andrew Blair-Stanek, Nils Holzenberger, and Benjamin~Van Durme. 2023.
\newblock \href {https://arxiv.org/abs/2302.06100} {Can gpt-3 perform statutory reasoning?}

\bibitem[{Chalkidis et~al.(2019)Chalkidis, Androutsopoulos, and Aletras}]{chalkidis2019neural}
Ilias Chalkidis, Ion Androutsopoulos, and Nikolaos Aletras. 2019.
\newblock Neural legal judgment prediction in english.
\newblock \emph{Association for Computational Linguistics (ACL)}.

\bibitem[{Chalkidis et~al.(2021)Chalkidis, Fergadiotis, Tsarapatsanis, Aletras, Androutsopoulos, and Malakasiotis}]{chalkidis-etal-2021-paragraph}
Ilias Chalkidis, Manos Fergadiotis, Dimitrios Tsarapatsanis, Nikolaos Aletras, Ion Androutsopoulos, and Prodromos Malakasiotis. 2021.
\newblock \href {https://doi.org/10.18653/v1/2021.naacl-main.22} {Paragraph-level rationale extraction through regularization: A case study on {E}uropean court of human rights cases}.
\newblock In \emph{Proceedings of the 2021 Conference of the North American Chapter of the Association for Computational Linguistics: Human Language Technologies}, pages 226--241, Online. Association for Computational Linguistics.

\bibitem[{Chen et~al.(2019)Chen, Cai, Dai, Dai, and Ding}]{chen-etal-2019-charge}
Huajie Chen, Deng Cai, Wei Dai, Zehui Dai, and Yadong Ding. 2019.
\newblock \href {https://doi.org/10.18653/v1/D19-1667} {Charge-based prison term prediction with deep gating network}.
\newblock In \emph{Proceedings of the 2019 Conference on Empirical Methods in Natural Language Processing and the 9th International Joint Conference on Natural Language Processing (EMNLP-IJCNLP)}, pages 6362--6367, Hong Kong, China. Association for Computational Linguistics.

\bibitem[{Kapoor et~al.(2022)Kapoor, Dhawan, Goel, T~H, Bhatnagar, Agrawal, Agrawal, Bhattacharya, Kumaraguru, and Modi}]{kapoor-etal-2022-hldc}
Arnav Kapoor, Mudit Dhawan, Anmol Goel, Arjun T~H, Akshala Bhatnagar, Vibhu Agrawal, Amul Agrawal, Arnab Bhattacharya, Ponnurangam Kumaraguru, and Ashutosh Modi. 2022.
\newblock \href {https://doi.org/10.18653/v1/2022.findings-acl.278} {{HLDC}: {H}indi legal documents corpus}.
\newblock In \emph{Findings of the Association for Computational Linguistics: ACL 2022}, pages 3521--3536, Dublin, Ireland. Association for Computational Linguistics.

\bibitem[{Katz et~al.(2023)Katz, Bommarito, Gao, and Arredondo}]{barExamGPT4}
Daniel~Martin Katz, Michael~James Bommarito, Shang Gao, and Pablo Arredondo. 2023.
\newblock \href {https://dx.doi.org/10.2139/ssrn.4389233} {Gpt-4 passes the bar exam}.

\bibitem[{Kingma and Ba(2014)}]{kingma2014adam}
Diederik~P Kingma and Jimmy Ba. 2014.
\newblock Adam: A method for stochastic optimization.
\newblock \emph{arXiv preprint arXiv:1412.6980}.

\bibitem[{Lage-Freitas et~al.(2022)Lage-Freitas, Allende-Cid, Santana, and Oliveira-Lage}]{lage2022predicting}
Andr{\'e} Lage-Freitas, H{\'e}ctor Allende-Cid, Orivaldo Santana, and L{\'\i}via Oliveira-Lage. 2022.
\newblock Predicting brazilian court decisions.
\newblock \emph{PeerJ Computer Science}, 8:e904.

\bibitem[{Lin(2004)}]{lin-2004-rouge}
Chin-Yew Lin. 2004.
\newblock \href {https://aclanthology.org/W04-1013} {{ROUGE}: A package for automatic evaluation of summaries}.
\newblock In \emph{Text Summarization Branches Out}, pages 74--81, Barcelona, Spain. Association for Computational Linguistics.

\bibitem[{Liu et~al.(2019)Liu, Ott, Goyal, Du, Joshi, Chen, Levy, Lewis, Zettlemoyer, and Stoyanov}]{liu2019roberta}
Yinhan Liu, Myle Ott, Naman Goyal, Jingfei Du, Mandar Joshi, Danqi Chen, Omer Levy, Mike Lewis, Luke Zettlemoyer, and Veselin Stoyanov. 2019.
\newblock Roberta: A robustly optimized bert pretraining approach.
\newblock \emph{arXiv preprint arXiv:1907.11692}.

\bibitem[{Long et~al.(2019)Long, Tu, Liu, and Sun}]{long2019automatic}
Shangbang Long, Cunchao Tu, Zhiyuan Liu, and Maosong Sun. 2019.
\newblock Automatic judgment prediction via legal reading comprehension.
\newblock In \emph{Chinese Computational Linguistics: 18th China National Conference, CCL 2019, Kunming, China, October 18--20, 2019, Proceedings 18}, pages 558--572. Springer.

\bibitem[{Malik et~al.(2021)Malik, Sanjay, Nigam, Ghosh, Guha, Bhattacharya, and Modi}]{malik-etal-2021-ildc}
Vijit Malik, Rishabh Sanjay, Shubham~Kumar Nigam, Kripabandhu Ghosh, Shouvik~Kumar Guha, Arnab Bhattacharya, and Ashutosh Modi. 2021.
\newblock \href {https://doi.org/10.18653/v1/2021.acl-long.313} {{ILDC} for {CJPE}: {I}ndian legal documents corpus for court judgment prediction and explanation}.
\newblock In \emph{Proceedings of the 59th Annual Meeting of the Association for Computational Linguistics and the 11th International Joint Conference on Natural Language Processing (Volume 1: Long Papers)}, pages 4046--4062, Online. Association for Computational Linguistics.

\bibitem[{Modi et~al.(2023)Modi, Kalamkar, Karn, Tiwari, Joshi, Tanikella, Guha, Malhan, and Raghavan}]{modi-etal-2023-semeval}
Ashutosh Modi, Prathamesh Kalamkar, Saurabh Karn, Aman Tiwari, Abhinav Joshi, Sai~Kiran Tanikella, Shouvik~Kumar Guha, Sachin Malhan, and Vivek Raghavan. 2023.
\newblock \href {https://doi.org/10.18653/v1/2023.semeval-1.318} {{S}em{E}val-2023 task 6: {L}egal{E}val - understanding legal texts}.
\newblock In \emph{Proceedings of the 17th International Workshop on Semantic Evaluation (SemEval-2023)}, pages 2362--2374, Toronto, Canada. Association for Computational Linguistics.

\bibitem[{Niklaus et~al.(2021)Niklaus, Chalkidis, and St{\"u}rmer}]{niklaus2021swiss}
Joel Niklaus, Ilias Chalkidis, and Matthias St{\"u}rmer. 2021.
\newblock Swiss-judgment-prediction: A multilingual legal judgment prediction benchmark.
\newblock \emph{arXiv preprint arXiv:2110.00806}.

\bibitem[{Papineni et~al.(2002)Papineni, Roukos, Ward, and Zhu}]{papineni-etal-2002-bleu}
Kishore Papineni, Salim Roukos, Todd Ward, and Wei-Jing Zhu. 2002.
\newblock \href {https://doi.org/10.3115/1073083.1073135} {{B}leu: a method for automatic evaluation of machine translation}.
\newblock In \emph{Proceedings of the 40th Annual Meeting of the Association for Computational Linguistics}, pages 311--318, Philadelphia, Pennsylvania, USA. Association for Computational Linguistics.

\bibitem[{Paul et~al.(2023)Paul, Mandal, Goyal, and Ghosh}]{paul-2022-pretraining}
Shounak Paul, Arpan Mandal, Pawan Goyal, and Saptarshi Ghosh. 2023.
\newblock \href {https://arxiv.org/abs/2209.06049} {Pre-trained language models for the legal domain: A case study on indian law}.
\newblock In \emph{Proceedings of 19th International Conference on Artificial Intelligence and Law - ICAIL 2023}.

\bibitem[{{\c{S}}ulea et~al.(2017){\c{S}}ulea, Zampieri, Malmasi, Vela, Dinu, and van Genabith}]{csulea2017exploring}
Octavia-Maria {\c{S}}ulea, Marcos Zampieri, Shervin Malmasi, Mihaela Vela, Liviu~P Dinu, and Josef van Genabith. 2017.
\newblock Exploring the use of text classification in the legal domain.

\bibitem[{Team et~al.(2023)Team, Anil, Borgeaud, Wu, Alayrac, Yu, Soricut, Schalkwyk, Dai, Hauth et~al.}]{team2023gemini}
Gemini Team, Rohan Anil, Sebastian Borgeaud, Yonghui Wu, Jean-Baptiste Alayrac, Jiahui Yu, Radu Soricut, Johan Schalkwyk, Andrew~M Dai, Anja Hauth, et~al. 2023.
\newblock Gemini: a family of highly capable multimodal models.
\newblock \emph{arXiv preprint arXiv:2312.11805}.

\bibitem[{Touvron et~al.(2023)Touvron, Martin, Stone, Albert, Almahairi, Babaei, Bashlykov, Batra, Bhargava, Bhosale et~al.}]{touvron2023llama}
Hugo Touvron, Louis Martin, Kevin Stone, Peter Albert, Amjad Almahairi, Yasmine Babaei, Nikolay Bashlykov, Soumya Batra, Prajjwal Bhargava, Shruti Bhosale, et~al. 2023.
\newblock Llama 2: Open foundation and fine-tuned chat models.
\newblock \emph{arXiv preprint arXiv:2307.09288}.

\bibitem[{Tunstall et~al.(2023)Tunstall, Beeching, Lambert, Rajani, Rasul, Belkada, Huang, von Werra, Fourrier, Habib et~al.}]{tunstall2023zephyr}
Lewis Tunstall, Edward Beeching, Nathan Lambert, Nazneen Rajani, Kashif Rasul, Younes Belkada, Shengyi Huang, Leandro von Werra, Cl{\'e}mentine Fourrier, Nathan Habib, et~al. 2023.
\newblock Zephyr: Direct distillation of lm alignment.
\newblock \emph{arXiv preprint arXiv:2310.16944}.

\bibitem[{Vasilyev et~al.(2020)Vasilyev, Dharnidharka, and Bohannon}]{blanc}
Oleg~V. Vasilyev, Vedant Dharnidharka, and John Bohannon. 2020.
\newblock \href {http://arxiv.org/abs/2002.09836} {Fill in the {BLANC:} human-free quality estimation of document summaries}.
\newblock \emph{CoRR}, abs/2002.09836.

\bibitem[{Vats et~al.(2023)Vats, Zope, De, Sharma, Bhattacharya, Nigam, Guha, Rudra, and Ghosh}]{vats-etal-2023-llms}
Shaurya Vats, Atharva Zope, Somsubhra De, Anurag Sharma, Upal Bhattacharya, Shubham Nigam, Shouvik Guha, Koustav Rudra, and Kripabandhu Ghosh. 2023.
\newblock \href {https://doi.org/10.18653/v1/2023.findings-emnlp.831} {{LLM}s {--} the good, the bad or the indispensable?: A use case on legal statute prediction and legal judgment prediction on {I}ndian court cases}.
\newblock In \emph{Findings of the Association for Computational Linguistics: EMNLP 2023}, pages 12451--12474, Singapore. Association for Computational Linguistics.

\bibitem[{Xiao et~al.(2018)Xiao, Zhong, Guo, Tu, Liu, Sun, Feng, Han, Hu, Wang et~al.}]{xiao2018cail2018}
Chaojun Xiao, Haoxi Zhong, Zhipeng Guo, Cunchao Tu, Zhiyuan Liu, Maosong Sun, Yansong Feng, Xianpei Han, Zhen Hu, Heng Wang, et~al. 2018.
\newblock Cail2018: A large-scale legal dataset for judgment prediction.
\newblock \emph{arXiv preprint arXiv:1807.02478}.

\bibitem[{Xu et~al.(2020)Xu, Wang, Chen, Pan, Wang, and Zhao}]{xu-etal-2020-distinguish}
Nuo Xu, Pinghui Wang, Long Chen, Li~Pan, Xiaoyan Wang, and Junzhou Zhao. 2020.
\newblock \href {https://doi.org/10.18653/v1/2020.acl-main.280} {Distinguish confusing law articles for legal judgment prediction}.
\newblock In \emph{Proceedings of the 58th Annual Meeting of the Association for Computational Linguistics}, pages 3086--3095, Online. Association for Computational Linguistics.

\bibitem[{Yang et~al.(2019{\natexlab{a}})Yang, Jia, Zhou, and Luo}]{10.5555/3367471.3367609}
Wenmian Yang, Weijia Jia, Xiaojie Zhou, and Yutao Luo. 2019{\natexlab{a}}.
\newblock Legal judgment prediction via multi-perspective bi-feedback network.
\newblock In \emph{Proceedings of the 28th International Joint Conference on Artificial Intelligence}, IJCAI'19, page 4085–4091. AAAI Press.

\bibitem[{Yang et~al.(2019{\natexlab{b}})Yang, Dai, Yang, Carbonell, Salakhutdinov, and Le}]{yang2019xlnet}
Zhilin Yang, Zihang Dai, Yiming Yang, Jaime Carbonell, Russ~R Salakhutdinov, and Quoc~V Le. 2019{\natexlab{b}}.
\newblock Xlnet: Generalized autoregressive pretraining for language understanding.
\newblock \emph{Advances in neural information processing systems}, 32.

\bibitem[{Zhang et~al.(2020)Zhang, Kishore, Wu, Weinberger, and Artzi}]{BERTScore}
Tianyi Zhang, Varsha Kishore, Felix Wu, Kilian~Q. Weinberger, and Yoav Artzi. 2020.
\newblock \href {https://openreview.net/forum?id=SkeHuCVFDr} {Bertscore: Evaluating text generation with {BERT}}.
\newblock In \emph{8th International Conference on Learning Representations, {ICLR} 2020, Addis Ababa, Ethiopia, April 26-30, 2020}. OpenReview.net.

\bibitem[{Zhong et~al.(2020)Zhong, Wang, Tu, Zhang, Liu, and Sun}]{zhong2020iteratively}
Haoxi Zhong, Yuzhong Wang, Cunchao Tu, Tianyang Zhang, Zhiyuan Liu, and Maosong Sun. 2020.
\newblock Iteratively questioning and answering for interpretable legal judgment prediction.
\newblock In \emph{Proceedings of the AAAI Conference on Artificial Intelligence}, volume~34, pages 1250--1257.

\end{thebibliography}
\bibliographystyle{acl_natbib}
\newpage
\appendix
\newpage
\section{Experimental Setup and Hyper-parameters}
\label{sec:Experimental-setup}

Our experimental setup was designed to optimize the performance of instruction fine-tuning on LLMs and to accurately assess their capabilities in legal judgment prediction and explanation tasks. We utilized two cores of NVIDIA A100-PCIE-40GB with 126GB RAM of 32 cores for instruction fine-tuning, ensuring powerful computational resources for processing and model training. In addition to the dedicated hardware, we employed a Google Colab Pro subscription having A100 Hardware accelerator for conducting inference and other experiments. This platform provided us with the necessary flexibility and scalability for our extensive experimentation.

Regarding the model training specifics, we fine-tuned the LLMs for 5 epochs. This duration was chosen to balance between adequately training the models on our \texttt{PredEx} dataset and preventing overfitting. During our experiments, we encountered a common issue with generative models – the tendency to hallucinate and repeat sentences. To address this, we implemented a post-processing step after inference. This step involved selecting the first occurrences of the decision and explanation parts from the model outputs and omitting any subsequent repetitions. This approach helped us refine the output quality, ensuring the results to be coherent and concise.

However, it is important to note that certain LLMs did not yield inference results in some cases. In such instances, we excluded those cases from our evaluation process. This decision was made to maintain the integrity and accuracy of our experimental findings, as including non-inferential results could have skewed our overall assessment of the models' performance.

Overall, our experimental setup was carefully crafted to provide a robust and reliable framework for evaluating the efficacy of instruction-tuned LLMs in the context of legal judgment prediction and explanation.


\section{Hallucination Examples}
\label{sec:Hallucination-example}

\subsection{Pre-trained vs Fine-tuned}
In the Appendix, we conduct a thorough comparison between pre-trained and fine-tuned models to demonstrate the reduction of hallucinations through our fine-tuning methods. Table \ref{tab:ca_analysis} in the Appendix presents this comparative analysis, illustrating the impact of instruction-tuning on legal judgment prediction with explanation tasks in our PredEx dataset. This section specifically focuses on how instruction-tuning can mitigate the issues of inaccuracies often found in the responses generated by pre-trained Large Language Models (LLMs).

\subsection{Zephyr-7b-beta and Gemini pro}
For a practical illustration of hallucination issues, Table \ref{zephyr-gemini-hallucination-example} in the Appendix provides examples from the Zephyr-7b-beta and Gemini Pro models. Notably, Zephyr-7b-beta sometimes produced contradictory outcomes (`accepted' and `rejected') for the same case with identical explanations. Additionally, an instance was observed where Gemini Pro produced a decision in Chinese, despite the prompt and the rest of the proceedings being in English. These examples highlight the critical need for model training and tuning specific to the dataset and task to ensure the reliability and accuracy of AI-generated legal predictions and explanations.


\begin{table*}[h]
\centering
\tiny
\resizebox{\textwidth}{!}{%
\begin{tabular}{|c|l|}
\hline

\textbf{Model} &
  \multicolumn{1}{c|}{\textbf{Decision and Explanation}} \\ \hline
 
Ground Truth &
  { \begin{tabular}[c]{@{}l@{}}Decsion: Accepted. Explanation: Insofar as first issue is concerned that need \\ not detain us for long as we find that the treatment given thereto by the High \\ Court is without any blemish.................................................The High Court \\ has held that it provides for only one situation where the time period can be \\ stretched, viz. where the assessment is held up (after due service of notice) \\ due to a stay order by a court of law. That covers only one eventuality, namely \\ assessment proceedings. The fact that the other periods have not been \\ mentioned leads to the inference that the delay in issuance of notice, or \\ delays in other situations have been intentionally left out, and the \\ consequence of such imperatively framed time period being breached, \\ result in invalidity of the act.\end{tabular}} \\ \hline
Zephyr-7b-beta &
  { \begin{tabular}[c]{@{}l@{}}{[}1, The appeal is likely to be allowed.{]}\\ \\ Explanation: The High Court has held that the notice issued for revising the \\ assessment list was not valid for the assessment year 1997-98 as it was \\ issued after the expiry of the assessment year. However, the notice was \\ received during the assessment year 1998-99. The Court has clarified that \\ the notice could not have been used to revise the assessment list for the \\ year 1997-98, but it was valid for the year 1998-99. Therefore, the Court is \\ likely to allow the appeal as the assessment list for the year 1998-99 was \\ validly revised.\\ \\ {[}0, The appeal is not likely to be allowed.{]}\\ \\ Explanation: The High Court has held that the notice issued for revising the \\ assessment list was not valid for the assessment year 1997-98 as it was \\ issued after the expiry of the assessment year. However, the notice was \\ received during the assessment year 1998-99. The Court has clarified that \\ the notice could not have been used to revise the assessment list for the \\ year 1997-98, but it was validly used to revise the assessment list for the \\ year 1998-99. Therefore, the appeal is not likely to be allowed as the issue \\ at hand is related to the assessment year 1997-98, for which the notice was \\ not valid.\end{tabular}} \\ \hline

Ground Truth &
  { \begin{tabular}[c]{@{}l@{}}Decision: Rejected \\ Explanation: At the outset we would like to say that we do not find any \\ substance in the contention of learned counsel for the appellant that there \\ has been any confusion of the issue or that any wrong approach has been \\ adopted by the taxing authorities, the Tribunal or the High Court.................\\ .............no deduction in respect of interest credited by the assessee to the \\ account of the college over the said sum can be allowedIn the circumstances, \\ in our view, the High Court rightly answered the questions referred to it \\ against the assessee in both the references.\end{tabular}} \\ \hline
Gemini pro &
  {\begin{tabular}[c]{@{}l@{}}Decision: \textcolor{red}{Chinese Character} \\Explanation: The High Court held that such a view was clearly sustainable \\ and observed that it is not the requirement under s. 27(iii) that the assessee \\ must further show that the borrowing of the capital was necessary for the \\ business so that if at the time of the borrowing the assessee has sufficient \\ amount of its own the deduction could not be allowed and the High Court\\  further took the view that in deciding whether a claim of interest on borrowing \\ can be allowed the fact that the assessee had ample resources its disposal \\ and need not have borrowed was not a relevant matter for consideration.\end{tabular}} \\ \hline
\end{tabular}%
}
\caption{We observe both `accept' and `reject' judgments in the response generated by Zephyr-7B-beta with the same explanation and opposite conclusion and repetitive responses. On the other hand, Gemini Pro gives responses in the Chinese language instead of English language in a few instances.}
\label{zephyr-gemini-hallucination-example}
\end{table*}

\begin{table*}[t]
\centering
\resizebox{\textwidth}{!}{%
\begin{tabular}{|c|l|l|ll}
\cline{1-3}
\textbf{Model}                                                                          & \multicolumn{1}{c|}{\textbf{Decision and Explanantion}}                  & \multicolumn{1}{c|}{\textbf{Explanation of error}}                      &  &  \\ \cline{1-3}
Ground Truth                                                                                & {\begin{tabular}[c]{@{}l@{}}Case decision: Accepted\\ Explanation:Thus, going by the principles governing the matter, propounded \\ by this Court there cannot be any doubt that the rights of the appellant-bank \\ over the pawned sugar had precedence over the claims of the Cane \\ Commissioner and that of the workmen. The High Court was, therefore, in \\ error in passing an interim order to pay parts of the proceeds to the Cane \\ Commissioner and to the Labour Commissioner for disbursal to the cane \\ growers and to the employees. There is no dispute that the sugar was \\ pledged with the appellant bank for securing a loan of the first respondent \\ and the loan had not been repaid. The goods were forcibly taken possession\\  of at the instance of the revenue recovery authority from the custody of the \\ pawnee, the appellant-bank. In view of the fact that the goods were validly \\ pawned to the appellant bank, the rights of the appellant-bank as pawnee \\ cannot be affected by the orders of the Cane Commissioner or the demands\\  made by him or the demands made on behalf of the workmen. Both the \\ Cane Commissioner and the workmen in the absence of a liquidation, stand\\  only as unsecured creditors and their rights cannot prevail over the rights of\\  the pawnee of the goods.19. We are also of the view that pending the writ \\ appeals, the High Court ought not to have passed such an interim order of \\ consequence especially in the light of the legal principles settled by this \\ Court. The order of the High Court, therefore, cannot be sustained and calls\\  for interference.\end{tabular}} & &  &  \\ \cline{1-3}
Llama-2-7B                                                                                  & { \begin{tabular}[c]{@{}l@{}}Case decision: Rejected\\ Explanation:\\ \\ The case proceedings provide several significant sentences that indicate the\\  outcome of the case will be rejected. Firstly, the court notes that the rights of\\  a pawnee over the pawned goods have precedence over the claims of a \\ secured creditor, even in the absence of a liquidation (para 15). This \\ suggests that the appellant bank's rights as a pawnee cannot be affected by\\  the orders of the Cane Commissioner or the demands made by him, even\\  though the goods were taken possession of at the instance of the revenue \\ recovery authority.\\ \\ Furthermore, the court observes that the Cane Commissioner and the\\  workmen stand only as unsecured creditors in the absence of a liquidation,\\  and their rights cannot prevail over the rights of the pawnee of the goods \\ (para 16). This implies that the appellant bank's rights as a pawnee are\\  superior to those of the Cane Commissioner and the workmen, and cannot \\ be ignored or set aside in favor of the latter.\\ \\ Finally, the court notes that the High Court's interim order was passed in \\ error, as it failed to recognize the appellant bank's rights as a pawnee \\ (para 18). This suggests that the High Court's decision will be reversed, and\\  the appellant bank's rights will be upheld.\\ \\ Based on these significant sentences in the case proceedings, it can be \\ concluded that the case will be rejected, and the appellant bank's rights as a\\  pawnee will be upheld.\end{tabular}}                                                                                & \begin{tabular}[c]{@{}l@{}}The Vanilla model incorrectly \\ categorises the sugarcane growers and\\  the workmen, represented by the Cane\\  commissioner and the Labour \\ Commissioner respectively, as "secured\\  creditors" when in actuality they are \\ "unsecured creditors" as already \\ established in the case text. This \\ hallucination is all the more apparent \\ as the model correctly identifies the \\ parties as "unsecured" creditors a few \\ lines later. Furthermore, the model \\ incorrectly places references to \\ different paragraphs of the case text,\\  i.e., para 15 and para 16. as the \\ paragraphs in question do not reflect\\  the text as purported by the model in \\ its output.\end{tabular} &  &  \\ \cline{1-3}
\begin{tabular}[c]{@{}c@{}}Llama-2-7B \\ Instruction-tuning\\ on prediction \\ with explanation\\ task\end{tabular} & \begin{tabular}[c]{@{}l@{}}Case decision: Accepted\\ Explanation:\\ \\ 18. Thus, going by the principles governing the matter, propounded by this \\ Court there cannot be any doubt that the rights of the appellant-bank over the\\  pawned sugar had precedence over the claims of the Cane Commissioner\\  and that of the workmen. The High Court was, therefore, in error in passing \\ an interim order to pay parts of the proceeds to the Cane Commissioner and\\  to the Labour Commissioner for disbursal to the cane growers and to the \\ employees. There is no dispute that the sugar was pledged with the appellant\\  bank for securing a loan of the first respondent and the loan had not been \\ repaid. The goods were forcibly taken possession of at the instance of the \\ revenue recovery authority from the custody of the pawnee, the \\ appellant-bank. In view of the fact that the goods were validly pawned to the\\  appellant bank, the rights of the appellant-bank as pawnee cannot be\\  affected by the orders of the Cane Commissioner or the demands made by\\  him or the demands made on behalf of the workmen. \textcolor{blue}{Both the Cane}\\  \textcolor{blue}{Commissioner and the workmen in the absence of a liquidation, stand} \\ \textcolor{blue}{only as unsecured creditors and their rights cannot prevail over the} \\ \textcolor{blue}{rights of the pawnee of the goods.}19. We are also of the view that pending\\  the writ appeals, the High Court ought not to have passed such an interim\\  order of consequence, especially in the light of the legal principles settled by\\  this Court. The order of the High Court, therefore, cannot be sustained and \\ calls for interference.\end{tabular}                                      & \begin{tabular}[c]{@{}l@{}}The Finetuned model accurately \\ categorises the workmen and the \\ sugarcane growers as "unsecured \\ creditors" and correctly determines that\\  the right of the pawnee (Appellant\\  Bank) will have precedence over their \\ rights to recompensation. Furthermore,\\  the finetuned model also accurately\\  states if the liquidation of the company\\  had been put into motion, the workmen\\  would THEN ONLY be considered \\ "secured creditors" in pari-passu with \\ other secured creditors.\end{tabular}&  &  \\ \cline{1-3}
\end{tabular}%
}
\caption{Comparative analysis of responses generated by Pretrained Llama-2-7B and Instruction Finetuned Llama-2-7B.}
\label{tab:ca_analysis}
\end{table*}
\begin{figure*}[h]
    \centering

    \begin{tcolorbox}[colframe=black, colback=white, boxrule=1pt, arc=0mm, outer arc=0mm]
    \includegraphics[trim=0.8cm 0.8cm 0.9cm 0.6cm, clip, scale=1]{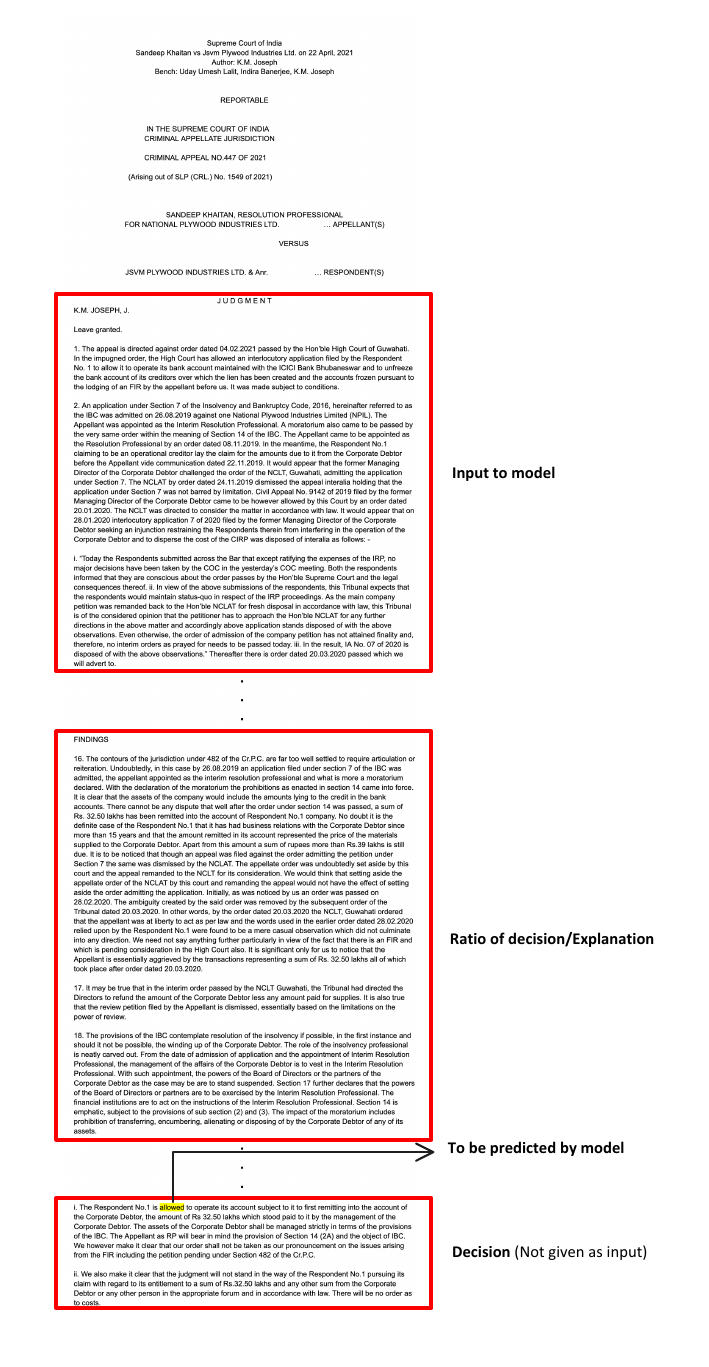}

    \end{tcolorbox}
    \caption{Annotated Example of Judicial Reasoning Extraction.}
    \label{fig:annotated-example}
\end{figure*}



\begin{table*}[t]
\centering
\resizebox{\textwidth}{!}{%
\begin{tabular}{|
>{}l |}
\hline
\textbf{CASE NO:} \\ \hline
Appeal (civil) 3499-3500 of 2007 \\ \hline
\textbf{PETITIONER:} \\ \hline
CENTRAL BANK OF INDIA \\ \hline
\textbf{RESPONDENT:} \\ \hline
SIRIGUPPA SUGARS \& CHEMICALS LTD. \& ORS \\ \hline
\textbf{DATE OF JUDGMENT:} \\ \hline
07/08/2007 \\ \hline
{ \textbf{BENCH:}} \\ \hline
{ TARUN CHATTERJEE \& P.K. BALASUBRAMANYAN} \\ \hline
\textbf{CASE TEXT:} \\ \hline
{ \begin{tabular}[c]{@{}l@{}}...These appeals challenge the interim order passed by the Division Bench of the High Court in a pending writ appeal, directing \\ disbursement of certain amounts realised on sale of stocks of sugar, owned by the first respondent company held under\\  pledge by the appellant--bank. The Labour Commissioner had passed an order under \textcolor{blue}{Section 33(c)} of the Industrial \\ Disputes Act against the first respondent company in respect of the dues to the workmen. The same was challenged by the \\ first respondent in the writ petition as also by others...\\ \\  ...In \textcolor{magenta}{Giles vs. Grover (1832 (131) ER 563 : 9 Bing 128)} it has been held that the Crown has no precedence over a pledgee of\\  goods. In Bank of Bihar vs. State of Bihar (supra) the principle has been recognised by this Court holding that the rights of the\\  pawnee who has parted with...\\ \\ ...There is no difference between the common law of England and the law with regard to pledge as codified. Under \\ \textcolor{blue}{Section 172} a pledge is a bailment of the goods as security for payment of a debt or  performance of a promise. \textcolor{blue}{Section 173}\\  entitles a pawnee to retain the goods pledged as security for payment of a debt and\\  under \textcolor{blue}{Section 175} he is entitled to receive from the pawner any extraordinary expenses he incurs for the preservation of the\\  goods pledged with him... \\ \\ ...In \textcolor{magenta}{State of M.P. vs. Jaura Sugar Mills Ltd. And others} (supra) dealing with the Madhya Pradesh\\ Sugar Cane (Regulation and Supply) Act, it was only held that the Cane Commissioner having power to compel the cane\\  growers to supply cane to the factory, has incidental power and is duty bound to ensure payment of the price of the\\  sugarcane supplied by the sugarcane growers...\end{tabular}} \\ \hline
\textbf{JUDGEMENT:} \\ \hline
{ \begin{tabular}[c]{@{}l@{}}...We, therefore, \textcolor{ForestGreen}{allow these appeals} and set aside the impugned order of the High Court, directing payment out of parts of the\\  sale proceeds to the Labour Commissioner and to the Cane Commissioner. \textcolor{ForestGreen}{We hold that the appellant as the pawnee, is} \\ \textcolor{ForestGreen}{entitled to the amount in satisfaction of its debt to secure which, the goods had been pawned and to appropriate the} \\ \textcolor{ForestGreen}{sale proceeds towards the debt due and only if there is surplus...}\end{tabular}} \\ \hline
\end{tabular}%
}
\caption{Example of Indian Case Structure. Sections referenced are highlighted in blue, previous judgments cited are in magenta, and the final decision is indicated in green.}
\label{case-example}
\end{table*}

\begin{table*}[h]
    \centering
    \begin{tabular}{|p{0.8\textwidth}|}
    \hline
{\bf Template 1 (prediction + explanation)}\\
\hline
   
{\bf prompt} = f``````Task: Given a Supreme Court of India case proceeding enclosed in angle brackets $<$ $>$, your task is to predict the decision of the case (with respect to the appelant) and provide an explaination for the decision.\\

{\bf Prediction}: Given a case proceeding, the task is to predict the decision 0 or 1, where the label 1 corresponds to the acceptance of the appeal/petition of the appellant/petitioner and the label 0 corresponds to the rejection of the appeal/petition of the appellant/petitioner,  Explanation: The task is to explain how you arrived at the decision by predicting important sentences that lead to the decision. \\

{\bf Context}: Answer in a consistent style as shown in the following two examples: \\

  {\bf case\_proceeding}: \# case\_proceeding example 1\\

  {\bf Prediction}: \# example 1 prediction \\

  {\bf Explanation}: \# example 1 explanation\\

  {\bf case\_proceeding}: \# case\_proceeding example 2\\

 {\bf  Prediction}: \# example 2 prediction\\

 {\bf Explanation}: \# example 2 explanation\\

{\bf Instructions}: Learn from the above given two examples and perform the task for the following case proceeding. \\

case\_proceeding: $<$\{case\_proceeding\}$>$\\

Format your output in list format: [prediction, explanation]''''''\\

\hline
    {\bf Template 2 (prediction only)}\\
    \hline
   
{\bf prompt} = f``````Task: Given a Supreme Court of India case proceeding enclosed in angle brackets $<$ $>$, your task is to predict the decision of the case (with respect to the appellant).\\

{\bf Prediction}: Given a case proceeding, the task is to predict the decision 0 or 1, where the label 1 corresponds to the acceptance of the appeal/petition of the appellant/petitioner and the label 0 corresponds to the rejection of the appeal/petition of the appellant/petitioner \\

{\bf Context}: Answer in a consistent style as shown in the following two examples: \\

  {\bf case\_proceeding}: \# case\_proceeding example 1\\

  {\bf Prediction}: \# example 1 prediction \\

  {\bf case\_proceeding}: \# case\_proceeding example 2\\

  {\bf Prediction}: \# example 2 prediction\\

{\bf Instructions}: Learn from the above given two examples and perform the task for the following case proceeding. \\

{\bf case\_proceeding}: $<$\{case\_proceeding\}$>$\\

Give the output predicted case decision as either 0 or 1.''''''\\

\hline

    \end{tabular}
    \caption{Prompts for Judgment Prediction taken from \cite{vats-etal-2023-llms}.}
    \label{tab:judgment_prediction_prompts_few}
\end{table*}

\begin{table*}[h]
    \centering
    \begin{tabular}{|p{0.8\textwidth}|}
    \hline
    {\bf Template 3 (prediction only)}\\
    \hline
   
{\bf prompt} = f``````
\#\#\# {\bf Instructions}: Analyze the case proceeding and predict whether the appeal/petition will be rejected (0) or accepted (1). \\

\#\#\# \textbf{Input}: $<$\{case\_proceeding\}$>$\\

\#\#\# Response:
  ''''''\\
\hline
    {\bf Template 4 (prediction with explanation)}\\
    \hline
   
{\bf prompt} = f``````
\#\#\# {\bf Instructions}: Analyze the case proceeding and predict whether the appeal/petition will be accepted (1) or rejected (0), and subsequently provide an explanation behind this prediction with important textual evidence from the case. \\

\#\#\# \textbf{Input}: $<$\{case\_proceeding\}$>$\\

\#\#\# Response:
  ''''''\\
\hline

    \end{tabular}
    \caption{Prompts for Judgment Prediction used for instruction fine-tuned models. Instructions were randomly chosen from Table \ref{Instruction-sets}.}
    \label{tab:judgment_prediction_prompts_zero}
\end{table*}

\begin{table*}[h]
\centering
\resizebox{0.95\textwidth}{!}{%
\tiny
\begin{tabular}{|cl|}
\hline
\multicolumn{2}{|c|}{\textbf{\textcolor{blue}{Instruction sets for Predicting the Decision}}} \\ \hline
\multicolumn{1}{|c|}{1} &
  Analyze the case proceeding and predict whether the appeal/petition will be accepted (1) or rejected (0). \\ \hline
\multicolumn{1}{|c|}{2} &
  \begin{tabular}[c]{@{}l@{}}Based on the information in the case proceeding, determine the likely outcome: acceptance (1) or \\ rejection (0) of the appellant/petitioner's case.\end{tabular} \\ \hline
\multicolumn{1}{|c|}{3} &
  Review the case details and predict the decision: will the court accept (1) or deny (0) the appeal/petition? \\ \hline
\multicolumn{1}{|c|}{4} &
  \begin{tabular}[c]{@{}l@{}}Considering the arguments and evidence in case proceeding, predict the verdict: is it more likely to be in \\ favor (1) or against (0) the appellant?\end{tabular} \\ \hline
\multicolumn{1}{|c|}{5} &
  \begin{tabular}[c]{@{}l@{}}Examine the details of the case proceeding and forecast if the appeal/petition stands a chance of being \\ upheld (1) or dismissed (0).\end{tabular} \\ \hline
\multicolumn{1}{|c|}{6} &
  \begin{tabular}[c]{@{}l@{}}Assess the case proceedings and provide a prediction: is the court likely to rule in favor of (1) or against (0)\\ the appellant/petitioner?\end{tabular} \\ \hline
\multicolumn{1}{|c|}{7} &
  \begin{tabular}[c]{@{}l@{}}Interpret the case information and speculate on the court's decision: acceptance (1) or rejection (0) of the \\ presented appeal.\end{tabular} \\ \hline
\multicolumn{1}{|c|}{8} &
  \begin{tabular}[c]{@{}l@{}}Given the specifics of the case proceeding, anticipate the court's ruling: will it favor (1) or oppose (0) the \\ appellant’s request?\end{tabular} \\ \hline
\multicolumn{1}{|c|}{9} &
  \begin{tabular}[c]{@{}l@{}}Scrutinize the evidence and arguments in the case proceeding to predict the court's decision: will the appeal\\  be granted (1) or denied (0)?\end{tabular} \\ \hline
\multicolumn{1}{|c|}{10} &
  \begin{tabular}[c]{@{}l@{}}Analyze the legal arguments presented and estimate the likelihood of the court accepting (1) or rejecting (0) \\ the petition.\end{tabular} \\ \hline
\multicolumn{1}{|c|}{11} &
  \begin{tabular}[c]{@{}l@{}}From the information provided in the case proceeding, infer whether the court's decision will be positive (1) \\ or negative (0) for the appellant.\end{tabular} \\ \hline
\multicolumn{1}{|c|}{12} &
  \begin{tabular}[c]{@{}l@{}}Evaluate the arguments and evidence in the case and predict the verdict: is an acceptance (1) or rejection\\ (0) of the appeal more probable?\end{tabular} \\ \hline
\multicolumn{1}{|c|}{13} &
  \begin{tabular}[c]{@{}l@{}}Delve into the case proceeding and predict the outcome: is the judgment expected to be in support (1) or \\ in denial (0) of the appeal?\end{tabular} \\ \hline
\multicolumn{1}{|c|}{14} &
  \begin{tabular}[c]{@{}l@{}}Using the case data, forecast whether the court is likely to side with (1) or against (0) the \\ appellant/petitioner.\end{tabular} \\ \hline
\multicolumn{1}{|c|}{15} &
  \begin{tabular}[c]{@{}l@{}}Examine the case narrative and anticipate the court's decision: will it result in an approval (1) or \\ disapproval (0) of the appeal?\end{tabular} \\ \hline
\multicolumn{1}{|c|}{16} &
  \begin{tabular}[c]{@{}l@{}}Based on the legal narrative and evidentiary details in the case proceeding, predict the court's stance: \\ favorable (1) or unfavorable (0) to the appellant.\end{tabular} \\ \hline
\multicolumn{2}{|c|}{\textbf{\textcolor{blue}{Instruction sets for Integrated Approach for Prediction and Explanation}}} \\ \hline
\multicolumn{1}{|c|}{1} &
  \begin{tabular}[c]{@{}l@{}}First, predict whether the appeal in case proceeding will be accepted (1) or not (0), and then explain the \\ decision by identifying crucial sentences from the document.\end{tabular} \\ \hline
\multicolumn{1}{|c|}{2} &
  \begin{tabular}[c]{@{}l@{}}Determine the likely decision of the case (acceptance (1) or rejection (0)) and follow up with an \\ explanation highlighting key sentences that support this prediction.\end{tabular} \\ \hline
\multicolumn{1}{|c|}{3} &
  \begin{tabular}[c]{@{}l@{}}Predict the outcome of the case proceeding (1 for acceptance, 0 for rejection) and subsequently provide an\\  explanation based on significant sentences in the proceeding.\end{tabular} \\ \hline
\multicolumn{1}{|c|}{4} &
  \begin{tabular}[c]{@{}l@{}}Evaluate the case proceeding to forecast the court's decision (1 for yes, 0 for no), and elucidate the \\ reasoning behind this prediction with important textual evidence from the case.\end{tabular} \\ \hline
\multicolumn{1}{|c|}{5} &
  \begin{tabular}[c]{@{}l@{}}Ascertain if the court will uphold (1) or dismiss (0) the appeal in the case proceeding, and then clarify \\ this prediction by discussing critical sentences from the text.\end{tabular} \\ \hline
\multicolumn{1}{|c|}{6} &
  \begin{tabular}[c]{@{}l@{}}Judge the probable resolution of the case (approval (1) or disapproval (0)), and elaborate on this forecast\\  by extracting and interpreting significant sentences from the proceeding.\end{tabular} \\ \hline
\multicolumn{1}{|c|}{7} &
  \begin{tabular}[c]{@{}l@{}}Forecast the likely verdict of the case (granting (1) or denying (0) the appeal) and then rationalize your \\ prediction by pinpointing and explaining pivotal sentences in the case document.\end{tabular} \\ \hline
\multicolumn{1}{|c|}{8} &
  \begin{tabular}[c]{@{}l@{}}Assess the case to predict the court's ruling (favorably (1) or unfavorably (0)), and then expound on \\ this prediction by highlighting and analyzing key textual elements from the proceeding.\end{tabular} \\ \hline
\multicolumn{1}{|c|}{9} &
  \begin{tabular}[c]{@{}l@{}}Decide if the appeal in the case proceeding is more likely to be successful (1) or unsuccessful (0), and \\ then justify your decision by focusing on essential sentences in the document.\end{tabular} \\ \hline
\multicolumn{1}{|c|}{10} &
  \begin{tabular}[c]{@{}l@{}}Conjecture the end result of the case (acceptance (1) or non-acceptance (0) of the appeal), followed by \\ a detailed explanation using crucial sentences from the case proceeding.\end{tabular} \\ \hline
\multicolumn{1}{|c|}{11} &
  \begin{tabular}[c]{@{}l@{}}Predict whether the case will result in an affirmative (1) or negative (0) decision for the appeal, and then \\ provide a thorough explanation using key sentences to support your prediction.\end{tabular} \\ \hline
\multicolumn{1}{|c|}{12} &
  \begin{tabular}[c]{@{}l@{}}Estimate the outcome of the case (positive (1) or negative (0) for the appellant) and then give a reasoned \\ explanation by examining important sentences within the case documentation.\end{tabular} \\ \hline
\multicolumn{1}{|c|}{13} &
  \begin{tabular}[c]{@{}l@{}}Project the court's decision (favor (1) or against (0) the appeal) based on the case proceeding, and \\ subsequently give an in-depth explanation by analyzing relevant sentences from the document.\end{tabular} \\ \hline
\multicolumn{1}{|c|}{14} &
  \begin{tabular}[c]{@{}l@{}}Make a prediction on the court's ruling (acceptance (1) or rejection (0) of the petition), and then dissect \\ the proceeding to provide a detailed explanation using key textual passages.\end{tabular} \\ \hline
\multicolumn{1}{|c|}{15} &
  \begin{tabular}[c]{@{}l@{}}Speculate on the likely judgment (yes (1) or no (0) to the appeal) and then delve into the case proceeding \\ to elucidate your prediction, focusing on critical sentences.\end{tabular} \\ \hline
\multicolumn{1}{|c|}{16} &
  \begin{tabular}[c]{@{}l@{}}Hypothesize the court's verdict (affirmation (1) or negation (0) of the appeal), and then clarify this \\ hypothesis by interpreting significant sentences from the case proceeding.\end{tabular} \\ \hline
\end{tabular}%
}
 \caption{Instruction Sets for Predicting Legal Decisions and Providing Explanations.}
\label{Instruction-sets}
\end{table*}

\end{document}